%% file: root.tex
\documentclass[letterpaper, 10 pt, conference]{ieeeconf}  \IEEEoverridecommandlockouts                              % This command is only needed if 
\usepackage[utf8]{inputenc}

\makeatletter
\let\NAT@parse\undefined
\makeatother
\usepackage[numbers, sort&compress]{natbib}

\usepackage{multicol}
\usepackage{amsmath}
\usepackage{balance}
\usepackage{amssymb}
\usepackage[hidelinks]{hyperref}
\usepackage{bbm}
\usepackage{dsfont}
\usepackage[table]{xcolor}% http://ctan.org/pkg/xcolor
\usepackage[font=scriptsize]{caption}
\usepackage{subcaption}
\usepackage{graphicx}
\usepackage{xcolor}
\usepackage{pifont}% http://ctan.org/pkg/pifont
\usepackage{booktabs}
\usepackage{multirow}
\usepackage{graphicx}
\usepackage{float}
\usepackage{algorithm}
\usepackage{algpseudocode}
\usepackage[all=normal,bibbreaks=tight,paragraphs=tight,floats=tight]{savetrees}

\title{\Large {\bf
ShaSTA-Fuse}: Camera-LiDAR Sensor Fusion to Model Shape and Spatio-Temporal Affinities for 3D Multi-Object Tracking}

\author{Tara Sadjadpour$^{1}$, Rares Ambrus$^{2}$, and Jeannette Bohg$^{1}$% <-this % stops a space
\thanks{*Toyota Research Institute provided funds to support this work.}% <-this % stops a space
\thanks{$^{1}$Stanford University,
        {\tt\small \{tsadja, bohg\}@stanford.edu}}%
\thanks{$^{2}$Toyota Research Institute,
        {\tt\small rares.ambrus@tri.global}}%
}

\begin{document}
\maketitle
\thispagestyle{empty}
\pagestyle{empty}

%%%%%%%%%%%%%%%%%%%%%%%%%%%%%%%%%%%%%%%%%%%%%%%%%%%%%%%%%%%%%%%%%%%%%%%%%%%%%%%%
% \balance
\input{sections/abstract.tex}
\IEEEpeerreviewmaketitle

\input{sections/introduction.tex}
\input{sections/related_work.tex}
\input{sections/background.tex}
\input{sections/approach.tex}
\input{sections/data_prep.tex}
\input{sections/experiment.tex}
\input{sections/conclusion.tex}

{
    \footnotesize
    \bibliographystyle{IEEEtran}
    \bibliography{IEEEabrv, refs.bib}
}

\end{document}

%% file: sections/abstract.tex
\begin{abstract}

3D multi-object tracking (MOT) is essential for an autonomous mobile agent to safely navigate a scene. In order to maximize the perception capabilities of the autonomous agent, we aim to develop a 3D MOT framework that fuses camera and LiDAR sensor information. Building on our prior LiDAR-only work~\cite{sadjadpour2022shasta} that models shape and spatio-temporal affinities for 3D MOT, we propose a novel camera-LiDAR fusion approach for learning affinities. At its core, this work proposes a fusion technique that generates a rich sensory signal incorporating information about depth and distant objects to enhance affinity estimation for improved data association, track lifecycle management, false-positive elimination, false-negative propagation, and track confidence score refinement. Our main contributions include a novel fusion approach for combining camera and LiDAR sensory signals to learn affinities, and a first-of-its-kind multimodal sequential track confidence refinement technique that fuses 2D and 3D detections. Additionally, we perform an ablative analysis on each fusion step to demonstrate the added benefits of incorporating the camera sensor, particular for small, distant objects that tend to suffer from the depth-sensing limits and sparsity of LiDAR sensors. In sum, our technique achieves state-of-the-art performance on the nuScenes benchmark amongst multimodal 3D MOT algorithms using CenterPoint detections.

\end{abstract}

%% file: sections/introduction.tex
\section{Introduction}

3D multi-object tracking is an essential component of any autonomous mobile agent's perception module, allowing the agent to perform safe, well-informed motion planning and navigation by localizing surrounding objects in 3D space and time. We address the problem of leveraging multimodal perception to accomplish 3D multi-object tracking for autonomous driving. In particular, we focus on camera-LiDAR sensor fusion due to the prevalence of both sensors on autonomous mobile agents, as well as the complementary nature of these two sensor modalities. Though LiDAR point clouds offer depth perception, they are limited by the LiDAR sensor's depth-sensing range and the sparse, unstructured nature of point clouds. On the other hand, cameras offer rich, structured signals and often capture distant objects, but they lack the depth information that is necessary for high-functioning perception systems. 

Though many 3D multi-object tracking algorithms focus on LiDAR-only perception, numerous recent works in this field focus on camera-LiDAR sensor fusion due to the increased advantages of using both sensors. In fact, the top 5 trackers for the nuScenes benchmark~\cite{caesar2020nuscenes} are all camera-LiDAR trackers. As a result, we aim to extend our previous LiDAR-only work ShaSTA~\cite{sadjadpour2022shasta} with ShaSTA-Fuse, which leverages the complementary nature of the camera and LiDAR sensors to benefit from their combined ability to perceive depth, distant objects, and rich sensory signals. 

The vast majority of state-of-the-art trackers rely on the \textit{tracking-by-detection} paradigm. In this model for approaching the 3D multi-object tracking problem, tracking algorithms use off-the-shelf, state-of-the-art 3D detectors. Though this is a practical choice due to the high computational cost for training detectors, it is often difficult to compare the quality of trackers when they use different detectors to achieve their results. This is particularly relevant, since the primary tracking accuracy metric called Average Multi-Object Tracking Accuracy (AMOTA) is strongly biased toward detection quality~\cite{weng2019baseline, luiten2021hota, wang2022camo}. Even though CenterPoint~\cite{yin2021center} detections tend to be the standard for LiDAR-only tracking~\cite{yin2021center, chiu2020probabilistic, sadjadpour2022shasta, liang2022neural, zaech2022learnable}, there is no clear standard in prior multimodal tracking works~\cite{chiu2021probabilistic, liu2022bevfusion, wang2022camo, kim2021eagermot, chen2017multi, willes2022intertrack, osep2017combined}. Nonetheless, due to the standardization of CenterPoint detections, we use them in this work. 

%For this reason, in our paper, we explore the effects of different detectors on multimodal tracking accuracy. 

Besides detection quality, the keys to successful 3D multi-object tracking are accurate data association, track lifecycle management, false-positive elimination, false-negative propagation, and track confidence score assignment. Data association is the process of matching existing tracks to incoming detections. When tracks and detections remain unmatched, there are a wide range of outcomes we must investigate to correctly handle them. First, unmatched tracks representing objects that are no longer in the scene must be terminated, while unmatched detections representing new objects in the scene must be initialized as newborn tracks. This is referred to as track lifecycle management. Even though these possibilities exist, unmatched tracks and detections can also be a residual effect of poor detection quality. Some unmatched tracks are not meant to be terminated, but rather, the detector has failed to capture a detection for the track due to difficult conditions such as occlusion, so we must interpolate that missing detection with false-negative propagation. Moreover, unmatched detections can be a result of false-positive detections that must be eliminated. Though tracking algorithms tend to focus on a subset of these problems, our prior work ShaSTA~\cite{sadjadpour2022shasta} is the first tracking algorithm that addresses all of these issues through an affinity-based framework. Thus, we aim to extend that approach in this work. Finally, track confidence scores indicate the quality of the track relative to other tracks in a given time step. ShaSTA~\cite{sadjadpour2022shasta} is the first-ever tracking algorithm that proposes a sequential track confidence refinement technique, rather than directly using detection confidence scores for tracks. This proves to work exceptionally well, and for this reason, we extend this technique to leverage camera and LiDAR information through a multimodal track confidence refinement. 

% Though many 3D multi-object tracking algorithms focus on LiDAR-only perception, including our own prior work ShaSTA which ranks 1st in overall AMOTA amongst LiDAR-only methods using CenterPoint detections, many recent works in this field focus on camera-LiDAR sensor fusion due to the increased advantages of using both sensors. In fact, the top 5 trackers on the nuScenes leaderboard are all camera-LiDAR trackers. As a result, we aim to extend our previous work ShaSTA with ShaSTA-Fuse, which leverages the complementary nature of the camera and LiDAR sensors. 

Amongst multimodal camera-LiDAR trackers, there are two general approaches for integrating camera information into the 3D multi-object tracking algorithm. The first is a late fusion technique that extracts features from images produced by camera sensors and combines them with LiDAR features~\cite{chiu2021probabilistic, wang2022camo, willes2022intertrack}, while the second uses abstracted information, such as 2D bounding boxes in the image plane, from 2D camera-based detections~\cite{kim2021eagermot, wang2022deepfusionmot, osep2017combined}. Though each direction has proved to be successful, this work differs from past approaches by leveraging both types of camera-based information. 

% Additionally, the vast majority of state-of-the-art trackers rely on the \textit{tracking-by-detection} paradigm. In this model for approaching the 3D multi-object tracking problem, tracking algorithms use off-the-shelf, state-of-the-art 3D detectors. Though this is a practical choice due to the high computational cost for training detectors, it is often difficult to compare the quality of trackers when they use different detectors to achieve their results. Even though CenterPoint detections tend to be the standard for LiDAR-only tracking, there is no clear standard in prior multimodal tracking works. For this reason, in our paper, we explore the effects of different detectors on tracking accuracy. 

% to maximize the depth perception of the LiDAR point cloud, while compensating for the sparsity and limited depth-sensing range of LiDAR using the camera sensor.

Thus, we propose a multimodal 3D multi-object tracking algorithm called \textit{ShaSTA-Fuse}, which \textbf{fuses} camera and LiDAR sensor signals to model \textbf{sha}pe and \textbf{s}patio-\textbf{t}emporal \textbf{a}ffinities. Our core contributions can be summarized in three main points: (i) we propose a novel late fusion approach for combining camera and LiDAR features in an affinity-based framework; (ii) we offer a simple yet effective multimodal sequential track confidence refinement technique that improves confidence score prediction by fusing 2D and 3D detections; (iii) we perform a thorough analysis on the most effective ways to fuse camera and LiDAR information. In sum, \textit{ShaSTA-Fuse} achieves state-of-the-art performance on the nuScenes tracking benchmark. 

%% file: sections/related_work.tex
\section{Related Work}\label{sec:related}

\subsection{2D Object Detection}
% \textcolor{red}{TODO: camera-only}

Image-based 2D object detection has rapidly advanced in recent years with the rise of deep learning. The vast majority of state-of-the-art 2D detectors use deep learning networks to extract features from input images in order to classify and localize objects of interest in the scene. There are two main branches of image-based object detectors: two-stage and one-stage detectors. 

The most well-known iterations of two-stage detectors are based on R-CNN~\cite{girshick2014rich}, which was followed by improvements known as Fast R-CNN~\cite{girshick2015fast}, Faster R-CNN~\cite{ren2015faster}, Cascade R-CNN~\cite{cai2018cascade}, and many more. In this paradigm, we begin with feature extraction followed by two-stage detection. In the first stage, the Region Proposal Network (RPN) proposes candidate object bounding boxes located in Regions of Interest (RoI). Then, in the second stage, features are extracted for each candidate bounding box using RoI Pooling and these features are passed into one or more Bounding Box Heads to further improve the initial bounding box prediction by regression on the object classification and bounding box.

Unlike two-stage detectors, one-stage detectors directly predict object classifications and bounding boxes from input images. The advent of this class of detectors came with the YOLO~\cite{redmon2016you} (You Only Look Once) detector, which was followed by improved iterations much like R-CNN~\cite{girshick2014rich}. 

The choice between these two approaches can be encapsulated as a speed-accuracy tradeoff. While the two-stage detectors maximize object classification and localization accuracy, the one-stage detectors achieve faster inference speed. 

Since the tracking task is largely dependent on detection quality, we opt for a two-stage detector in Cascade R-CNN~\cite{cai2018cascade} for our algorithm. Additionally, we leverage a Cascade R-CNN network that has been pre-trained on nuScenes image data to extract salient appearance features from our camera data, freezing the entire network except for the Bounding Box Head which we fine-tune. Past 3D multi-object tracking works have not characterized objects localized by their projected 3D detection locations using features extracted in the Bounding Box Head. 

\subsection{3D Object Detection}
% \textcolor{red}{TODO: multimodal vs LiDAR-only}

There is a vast array of approaches in the 3D object detection domain, which faces increased complexity compared to 2D object detection. For 3D object detection we see detectors of various modalities, including LiDAR-only, camera-LiDAR, and the lesser used camera-only. 

\subsubsection{LiDAR-Only 3D Detectors}

In the LiDAR-only domain, older works~\cite{wang2015voting, song2014sliding, song2016deep, chen2017multi, li20173d} hand-craft dense feature representations from the sparse LiDAR point cloud. Using these feature maps, the 3D detectors extend the image-based R-CNN~\cite{girshick2014rich} to 3D with an RPN followed by regression heads to improve the initial 3D bounding box estimates. Building on the momentum of works that learn point-wise features from point clouds using deep learning~\cite{qi2017pointnet, qi2017pointnet++}, VoxelNet~\cite{zhou2018voxelnet} was the first LiDAR-only 3D detector to unify feature extraction and bounding box prediction into an end-to-end learned, one-stage detection framework. Following this work, it became typical for 3D detectors to start with a deep learning-based point cloud encoder~\cite{qi2017pointnet,engelcke2017vote3deep,yang2018pixor} to extract an intermediary 3D feature grid or BEV feature map~\cite{lang2019pointpillars,yin2021center, wang2021object, wang2022camo}. Then, a decoder is applied to extract object attributes including classification, localization, and detection confidence score. In this work, we explore fusing information from 2D detections produced by Cascade R-CNN~\cite{cai2018cascade} with 3D detections.  We explore this fusion approach with the LiDAR-only detector CenterPoint~\cite{yin2021center}. Additionally, we leverage the pre-trained VoxelNet~\cite{zhou2018voxelnet} LiDAR backbone from~\cite{yin2021center} to extract an intermediate BEV feature map which we process for late fusion with our image-based appearance features.

%In one ablation study, we explore this fusion approach with the LiDAR-only detector~\cite{yin2021center}. Additionally, we leverage the pre-trained LiDAR backbone from~\cite{zhou2018voxelnet, yin2021center} to extract an intermediate BEV feature map which we process for late fusion with our image-based appearance features.

\subsubsection{Camera-Based 3D Detectors}
For improved results to compensate for the sparse signal from LiDAR sensors, the 3D detection community has developed increased interest in camera-LiDAR fusion. Proposal-level fusion methods~\cite{chen2017multi, bai2022transfusion, chen2022futr3d, nabati2011center, qi2018frustum, wang2019frustum} are object-centric techniques that try to fuse object proposals in either image or 3D space. The methods tend not to generalize to other tasks as well as point-level fusion approaches~\cite{vora2020pointpainting, wang2021pointaugmenting, yin2021multimodal, liu2022bevfusion}, which are both object-centric and geometric-centric. In this framework, background image features are painted onto foreground LiDAR points and LiDAR-based detection is completed on these augmented point clouds. In this work, we do not use such detectors, since there is no standard multimodal detector used in 3D multi-object tracking. For example, the best camera-LiDAR 3D detector to date is BEVFusion~\cite{liu2022bevfusion} which also produces state-of-the-art results on the nuScenes leaderboard~\cite{caesar2020nuscenes}. Based on how past works have performed, it is expected that the tracking results on the validation set are at least as accurate as those on the test set, but the authors' open-source code do not match their reported tracking results, with a significant drop in accuracy (AMOTA) of 3.7 points. The authors have not explained this discrepancy, and for this reason, we avoid using these detections altogether, since there is no fair way to assess our tracking results with these detections\footnote{\href{https://github.com/mit-han-lab/bevfusion/issues/256\#issuecomment-1378158369}{Link to GitHub issue about discrepancies in BEVFusion tracking results.} We tried one of the remedies suggested by a GitHub user to filter out low-scoring detections ($<0.025$), but this is still 1.8 points lower in AMOTA than the reported test set results. In the CAMO-MOT paper~\cite{wang2022camo}, the authors suggest that the BEVFusion detections used to generate the BEVFusion tracking results originate from ``a much more powerful detector" than the ones provided on the authors' GitHub page.}. 

%For example, when we use the open-source code provided by the authors of BEVFusion~\cite{liu2022bevfusion}, the best multimodal 3D detector, to replicate their tracking results, we found that their tracking accuracy (AMOTA) is 3.7 points lower in the validation set than the test set results they obtained on the nuScenes leaderboard~\cite{caesar2020nuscenes}. The authors are unable to explain for this discrepancy on their GitHub page, leaving others to speculate about potential ways to use their detections for tracking\footnote{\href{https://github.com/mit-han-lab/bevfusion/issues/256\#issuecomment-1378158369}{Link to GitHub issue about discrepancies in BEVFusion tracking results.}}. We tried one of the remedies suggested by a GitHub user to filter out low-scoring detections ($<0.025$), but this is still 1.8 points lower in AMOTA than the reported test set results. In the CAMO-MOT paper~\cite{wang2022camo}, the authors suggest that the BEVFusion detections used to generate the BEVFusion tracking results originate from ``a much more powerful detector" than the ones provided on the authors' GitHub page. For this reason, we avoid using these detections altogether, since there is no fair way to assess our tracking results with these detections. 

%BEVFusion~\cite{liu2022bevfusion} is a point-level fusion method, where geometric and semantic information are treated equally, and has been shown to generalize well to many tasks including 3D object detection, 3D multi-object tracking, and segmentation. %Thus, in this work we use the BEVFusion~\cite{liu2022bevfusion} detections. 

Finally, there is a line of work dedicated to camera-only 3D detectors in which 3D bounding boxes are estimated from 2D images. However, these detections yield poor accuracy due to poor depth estimation, and we do not conduct experiments with camera-only 3D detections.

\subsection{3D Multi-Object Tracking}
% s\textcolor{red}{TODO: Multimodal -- extract image features or not, reference LiDAR-only line of work but state this is not our focus}

In the 3D multi-object tracking domain, there have been significant efforts in LiDAR-only tracking~\cite{weng2019ab3dmot, chiu2020probabilistic, yin2021center, stearns2022spot, sadjadpour2022shasta, liang2022neural, zaech2022learnable}. Much of these works use Kalman Filters, explicit motion models, and various distance metrics, including Intersection Over Union (IoU) and Mahalanobis distance, to complete data association. Additionally, much of the algorithms use low-dimensional bounding boxes as abstracted object representations, since it is challenging to extract expressive features from sparse LiDAR point clouds. However, recent works in this field attempt to better leverage the LiDAR data for improved results. SpOT~\cite{stearns2022spot} maintains an extended temporal history beyond the Markov assumption with LiDAR point clouds accumulated over several key frames. This proves to be especially beneficial for objects that experience occlusion like pedestrians. Additionally, ShaSTA~\cite{sadjadpour2022shasta} leverages intermediate representations of LiDAR data from a detection network's LiDAR backbone for richer information about each object, while also modeling the global relationship of all the objects in a scene to improve data association. We build on this paradigm by leveraging intermediate representations of both camera and LiDAR data from 2D and 3D detection networks' backbones, respectively, with a novel late fusion approach to get richer object-level representations. 

In the multimodal 3D multi-object tracking domain for camera-LiDAR fusion, we see two main paradigms that are followed. On one hand, the tracking algorithms in this area process camera images to extract appearance features that are fused with LiDAR features for better tracking~\cite{chiu2021probabilistic, willes2022intertrack, wang2022camo, weng2020gnn3dmot}. On the other hand, an under-explored, yet effective method for developing such algorithms is by fusing information from 2D detection bounding boxes and 3D detection bounding boxes to improve tracking accuracy~\cite{kim2021eagermot, wang2022deepfusionmot}. The latter line of work often yields significantly higher accuracy gains for small object like bicycles without incurring the high computational expense of image feature extraction. In this work, we do not treat these two approaches as mutually exclusive. This work is the first to our knowledge that fuses camera and LiDAR features in a neural network to enrich object-level geometric representations, while fusing information from 2D and 3D detection bounding boxes for a novel multimodal track confidence refinement technique during track formation.

%% file: sections/background.tex
% \section{Background}

% % \textcolor{red}{TODO: discuss background on ShaSTA which is our own work we build off of}

% In this work, we build off of the LiDAR-only framework developed in ShaSTA. Thus, we offer background on the 

%% file: sections/approach.tex
\section{SHASTA-FUSE}
In this section, we begin by defining and offering intuition of the general affinity-based 3D multi-object tracking (MOT) framework developed in ShaSTA~\cite{sadjadpour2022shasta}. Then, we describe how we have effectively extended this affinity-based framework for multimodal 3D MOT. 
\subsection{Review of ShaSTA~\cite{sadjadpour2022shasta}: Affinity Estimation for LiDAR-Only 3D MOT}
Given a set of unordered point sets generated by a LiDAR sensor from the current and previous time steps $L_t$ and $L_{t-1}$, and 3D detection bounding boxes estimated by LiDAR-only 3D detectors from the current and previous time steps $B_t$ and $B_{t-1}$, one can define a function $f_i$ for each object class $i$ that learns an affinity matrix $A_t$: 
\begin{align}
    f_i(L_{t-1}, L_t, B_{t-1}, B_t) = A_t
\end{align}

For $N_{max}$ objects in a frame, the affinity matrix $A_t \in \mathbb{R}^{(N_{max}+2) \times (N_{max}+2)}$ not only matches existing tracks and current frame detections to complete data association, but also has augmented columns and rows to handle special case situations. The two additional columns are for dead-track (DT) and false-negative (FN) \textit{anchors}. This allows for existing tracks to match with either of these anchors to terminate the track or propagate it forward using its estimated velocity to compensate for an FN detection. Additionally, the two additional rows are for false-positive (FP) and newborn (NB) anchors, so extraneous detections can be eliminated as FPs or incoming detections can be initialized as NB tracks when they do not match with existing tracks. Instead of applying heuristics, we are matching to frame-specific anchors that have been learned by understanding the distribution of FPs, FNs, DTs, and NBs over the entire training set. For example, the FP anchor's bounding box center that ShaSTA learns for frame $t$ is a centroid of where most FP detections are in that frame. Therefore, we expect the current frame's FP detections to match with the FP anchor in the affinity matrix. 

While there is a unique one-to-one correspondence between bounding boxes in two consecutive frames, more than one current frame bounding box can match with the NB and FP anchors, and more than one existing track can match with the DT and FN anchors. To handle this situation, we define the forward and backward matching affinity matrices. Here, forward matching indicates that we want to find the best current frame match for each previous frame track, while backward matching refers to finding the best previous frame match for each current frame detection.

\input{figures/flowchart.tex}

Thus, we create a forward matching affinity matrix $A_{t,fm} \in \mathbb{R}^{N_{max}\times (N_{max}+2)}$ by removing the two row augmentations from $A_t$ and applying a row-wise softmax so that more than one previous frame track can have sufficient probability to match with the DT and FN anchors, respectively: 
\begin{align}
    A_{t,fm} =  \text{softmax}_\text{row}\left(A_{t,1}\right).
\end{align}

Using similar logic, we create a backward matching affinity matrix $A_{t,bm} \in \mathbb{R}^{(N_{max}+2) \times N_{max}}$ by removing the two column augmentations from $A$ and applying a column-wise softmax so that more than one current frame detection can have sufficient probability to match with the NB and FP anchors, respectively: 
\begin{align}
    A_{t,bm} =  \text{softmax}_\text{col}\left(A_{t,2}\right).
\end{align}

In order to learn the affinity matrix, we use supervised learning with the \textit{log affinity loss} defined in~\cite{sadjadpour2022shasta}. For each time step $t$, we define the ground-truth affinity matrices $A_{gt,fm}$ and $A_{gt,bm}$ for forward matching and backward matching, respectively. Thus, our loss function $\mathcal{L}$ is defined as follows:
\begin{align}
    \mathcal{L}_{fm} &= \frac{\sum_i \sum_j (A_{gt,fm} \odot -\log(A_{fm}))}{\sum_i \sum_j  A_{gt,fm}} \\ 
    \mathcal{L}_{bm} &= \frac{\sum_i \sum_j (A_{gt,bm} \odot -\log(A_{bm}))}{\sum_i \sum_j  A_{gt,bm}} \\
    \mathcal{L} &= \frac{1}{2} \Big(\mathcal{L}_{fm} + \mathcal{L}_{bm}\Big).
\end{align}

Finally, the information encoded in the affinity matrix prediction is used for sequential track confidence score refinement.

\subsection{Camera-LiDAR Sensor Fusion to Estimate Affinities for Multimodal 3D MOT}
We create ShaSTA-Fuse, the camera-LiDAR variation of ShaSTA, by integrating appearance cues into our affinity matrix estimation and using camera-based 2D detection information to create a multimodal track confidence refinement technique. These two additions constitute ShaSTA-Fuse, and we describe them below. A visualization of ShaSTA-Fuse can be found in Figure~\ref{fig:pipeline}.

\noindent \textbf{Appearance Cue Integration.} We integrate camera sensor information into the ShaSTA framework by extracting appearance cues from images. 

% For this section, we assume readers are familiar with~\cite{sadjadpour2022shasta}, where we define how we obtain the augmented bounding boxes and augmented shape descriptors $\hat{B}$ and $\hat{S}$, as well as the VoxelNet, bounding box, and shape residuals $R_v$, $R_b$, and $R_s$. 

The autonomous vehicles in the nuScenes benchmark are equipped with 6 camera sensors and 1 LiDAR sensor. Therefore, for each time step, we have 1 LiDAR point cloud and 6 images. Since the 3D detections are obtained with the LiDAR sensor in the world coordinate system, we map each 3D detection to all 6 image planes using each of the 6 camera transforms to see which camera the 3D detection corresponds to. If a projected 3D detection appears in more than one image plane, we choose the image in which it has the largest area with its projected bounding box defined by $b_{proj} := (x_{img}, y_{img}, w, h)$. For any time step $t$, we define the set of projected 3D detections as $B_{proj,t} \in \mathbb{R}^{N_{max} \times 4}$, where we have up to $N_{max}$ objects per frame. If there are fewer than $N_{max}$ objects in a frame, then we zero-pad the remaining entries, and if there are more than $N_{max}$ objects then we sample the top $N_{max}$ detections.

For each time step, we pass the 6 images into a pre-trained Cascade R-CNN detector~\cite{cai2018cascade}. After the feature map is created for each image, we forgo the RPN, and treat the projected 3D detections $B_{proj,t}$ as our RoIs. Thus, we perform RoI pooling with the projected 3D detections $B_{proj,t}$, and pass the extracted features corresponding to each RoI into our Bounding Box Head. Since the Bounding Box Head outputs bounding box predictions, we only use its initial layers that process the RoIs to get our appearance cues, which are $F_C$-dimensional. We do not freeze the Bounding Box Head layers, and allow them to be fine-tuned, since the RoIs we pass in are projections from the 3D detector and not Cascade R-CNN detections. For each pair of frames at the current and previous time steps $t$ and $t-1$, we obtain appearance cues $C_t \in \mathbb{R}^{N_{max} \times F_C}$ and $C_{t-1} \in \mathbb{R}^{N_{max} \times F_C}$, respectively.

Then, we create a learned appearance cue representation at time step $t$ for FP and NB anchors using the current frame appearance cues $C_t$ as follows, where each $\sigma$ represents an MLP: %$$b_{fp} = \sigma_{fp}^b(B_t)$$$ and $$$b_{nb} = \sigma_{nb}^b(B_t)$$.
\begin{align}
    c_{fp} &= \sigma_{fp}^c(C_t) \\
    c_{nb} &= \sigma_{nb}^c(C_t).
\end{align}

Similarly, we find learned appearance cue representations for FN and DT anchors using previous frame appearance cues $C_{t-1}$: %$b_{fn} = \sigma_{fn}^b(B_{t-1})$ and $b_{dt} = \sigma_{dt}^b(B_{t-1})$.
\begin{align}
    c_{fn} &= \sigma_{fn}^c(C_{t-1}) \\
    c_{dt} &= \sigma_{dt}^c(C_{t-1}).
\end{align}

We then concatenate $c_{fp} \in \mathbb{R}^{F_C}$ and $c_{nb} \in \mathbb{R}^{F_C}$ to $C_{t-1}$ to get $\hat{C}_{t-1} \in \mathbb{R}^{(N_{max}+2) \times F_C}$, as well as $c_{fn} \in \mathbb{R}^{F_C}$ and $c_{dt} \in \mathbb{R}^{F_C}$ to $C_{t}$ to get $\hat{C}_{t} \in \mathbb{R}^{(N_{max}+2) \times F_C}$.

We use $\hat{C}_t$ and $\hat{C}_{t-1}$ to obtain the learned appearance cue residual $R_c$ between the two frames. The residual measures the similarities between current and previous frames’ appearance cue representations. We expand $\hat{C}_{t-1}$ and $\hat{C}_t$ and concatenate them to get a matrix $\hat{C} \in \mathbb{R}^{(N_{max}+2)\times (N_{max}+2)\times 2F_C}$. Then, we obtain our residual $R_c \in \mathbb{R}^{(N_{max}+2)\times (N_{max}+2)}$ with an MLP: %$R_s = \sigma_r^s(\hat{S})$.
\begin{align}
    R_c = \sigma_r^c(\hat{C}).
\end{align}

In addition to the appearance cue residual $R_c$, we use the VoxelNet, bounding box, and shape residuals $R_v$, $R_b$, and $R_s$ that encode shape and spatio-temporal information from the LiDAR sensor. These residuals are formed in an analogous way to the appearance cue residual $R_c$. We refer readers to~\cite{sadjadpour2022shasta} for further details, as well as Figure~\ref{fig:pipeline} which shows how we incorporate all of the residuals.

The overall residual $R$ is a weighted sum of the VoxelNet, bounding box, and shape residuals we obtained in ShaSTA, known as $R_v$, $R_b$, and $R_s$, and the new appearance cue residual we obtain $R_c$. We concatenate $\hat{B}$, $\hat{S}$, and $\hat{C}$ to create our input $\hat{W} \in \mathbb{R}^{(N_{max}+2)\times(N_{max}+2)\times(2F_S + 2F_C +6)}$. Then, we pass this through an MLP $\sigma_\alpha$ to get the learned weights $\pmb{\alpha} \in \mathbb{R}^{(N_{max}+2)\times (N_{max}+2) \times 4}$ as follows: %$\pmb{\alpha} = \sigma_\alpha(\hat{W})$.
\begin{align}
    \pmb{\alpha} = \sigma_\alpha(\hat{W}).
\end{align}
We split $\pmb{\alpha}$ into matrices $\alpha_v$, $\alpha_b$, $\alpha_s$, $\alpha_c \in \mathbb{R}^{(N_{max}+2)\times (N_{max}+2)}$, and obtain our overall residual $R \in \mathbb{R}^{(N_{max}+2)\times (N_{max}+2)}$:
\begin{align}
    R = \alpha_v \odot R_v + \alpha_b \odot R_b + \alpha_s \odot R_s + \alpha_c \odot R_c.
\end{align}
Note that $\odot$ is the Hadamard product.

Given our overall residual, we have information about the pairwise similarities between current frame detections and previous frame tracks, as well as each of the four augmented anchors. In order to leverage these spatio-temporal and shape relationships to learn probabilities for matching the detections and tracks to each other or the anchors, we apply an MLP $\sigma_{aff}$ to get our overall affinity matrix $A_t$ for time step $t$:
\begin{align}
    A_t =  \sigma_{aff}(R).
\end{align}

\noindent \textbf{Multimodal Track Confidence Refinement.} We build off of the success of the LiDAR-only sequential track confidence refinement from~\cite{sadjadpour2022shasta} by integrating both camera and LiDAR information at this stage. 

For the sake of completeness, we remind readers that track confidence is a relative measure of track quality with respect to objects of the same class in a given time step. Just as was done in~\cite{sadjadpour2022shasta}, we treat track confidence score refinement as a sequential process that reflects changing environmental factors. For example, if a track has been maintained for an extended period of time but the object suddenly experiences occlusion, we do not let the confidence abruptly drop; rather, we take into account that history and use it to inform our confidence scores. 

In order to extend this technique to camera-LiDAR sensor fusion, we not only leverage infromation encoded in our learned affinity matrix just as in~\cite{sadjadpour2022shasta}, but we also use two detectors. For each time step $t$, we use the 2D detections from Cascade R-CNN~\cite{cai2018cascade} $B_{2D,t}$ and the projected 3D detections $B_{proj,t}$ that we created in the previous section. Additionally, for each object class, we choose an Intersection over Union (IoU) threshold $\tau_{iou}$. For each camera frame, we get the pairwise IoUs between $B_{2D,t}$ and $B_{proj,t}$. We greedily match 2D detections and projected 3D detections that achieve an IoU over the threshold $\tau_{iou}$. Thus, each detection from $B_{2D,t}$ is matched with at most one detection from $B_{proj,t}$, and visa versa. 

For each 3D detection $b_i^{(t)}$ at time step $t$ that matches with an existing track, we add the index $i$ to the set $I_{matched}$ for matched detections. Similarly, for each 3D detection $b_i^{(t)}$ that is designated as a newborn, we add the index $i$ to the set $I_{newborn}$ for detections that should be initialized as newborn tracks. For each 3D detection $b_i^{(t)}$, we have the probability that it matched to the false-positive anchor $P_{FP,i}$ which is learned in the affinity matrix $A_{bm}$. Note that the probability that a 3D detection is true-positive is defined as $P_{TP,i} = 1 - P_{FP,i}$. 

We then perform our multimodal track confidence refinement as shown in Algorithm~\ref{alg:mmcr}. Note that we fix $\beta_1$ to be a value slightly less than $\tau_{FP}$ to account for very uncertain detections that have been matched to tracks, and we effectively reduce the overall track confidence for tracks matching to such detections. However, if the matched detection has a very high probability of being true-positive, then the track confidence $c_{trk,i}^{(t)}$ becomes a weighted average between $c_{det,i}^{(t)}$ and $c_{trk,i}^{(t-1)}$. We set $\beta_1 = 0.5$ for all object classes, and $\beta_2=0.5$ for all object classes except for bicycles and cars ($\beta_2=0.4$), bus ($\beta_2 =0.7$), and trailer ($\beta_2=0.4$). Additionally, we define the function $\hat{f}$ as follows:
\begin{align}
    \hat{f}(x, IoU_i, \tau_{iou}) = \min\Bigg(\frac{IoU_i}{\tau_{iou}} \cdot x, 1\Bigg).
\end{align}
This function is meant to reward 3D detections that match with a 2D detection based on the IoU thresholding by increasing their confidence scores, increasing the weight they receive in the weighted average, and increasing their probability of being true-positive to avoid reducing the overall confidence of the tracks they match to. 

\begin{algorithm}
\caption{Multimodal Track Confidence Refinement}\label{alg:mmcr}
\begin{algorithmic}
\State $I_{newborn}$, $I_{matched}$, $\tau_{iou}$, $\beta_1$, $\beta_2$
\For{$i \in I_{matched}$}
\State $\beta_{2,i} \gets \beta_2$
\If {$IoU_i^{(t)} > \tau_{iou}$}
    \State $c_{det, i}^{(t)} \gets \hat{f}\Big(\max\Big(c_{3D-det, i}^{(t)}, c_{2D-det, i}^{(t)}\Big), IoU_i^{(t)}, \tau_{iou}\Big)$
    \State $\beta_{2,i} \gets \hat{f}\Big(\beta_2, IoU_i^{(t)}, \tau_{iou}\Big)$
    \State $P_{FP, i} \gets 1 - \hat{f}\Big(P_{TP, i}, IoU_i^{(t)}, \tau_{iou}\Big)$
\EndIf
\State $c_{trk, i}^{(t)} \gets \mathds{1}[P_{FP, i}^{(t)} < \beta_1]\beta_{2,i} c_{det, i}^{(t)} + (1-\beta_{2,i})c_{trk, i}^{(t-1)}$
\EndFor

\For{$i \in I_{newborn}$}
\State $\beta_{2,i} \gets \beta_2$
\If {$IoU_i^{(t)} > \tau_{iou}$}
    \State $c_{det, i}^{(t)} \gets \hat{f}\Big(\max\Big(c_{3D-det, i}^{(t)}, c_{2D-det, i}^{(t)}\Big), IoU_i^{(t)}, \tau_{iou}\Big)$
    \State $\beta_{2,i} \gets \hat{f}\Big(\beta_2, IoU_i^{(t)}, \tau_{iou}\Big)$
    \State $P_{FP, i} \gets 1 - \hat{f}\Big(P_{TP, i}, IoU_i^{(t)}, \tau_{iou}\Big)$
\EndIf
\State $c_{trk, i}^{(t)} \gets \mathds{1}[P_{FP, i}^{(t)} < \beta_1]\beta_{2,i} c_{det, i}^{(t)}$
\EndFor
\end{algorithmic}
\end{algorithm}

%% file: figures/flowchart.tex
\begin{figure*}[ht!]
\centering
\includegraphics[scale=0.12]{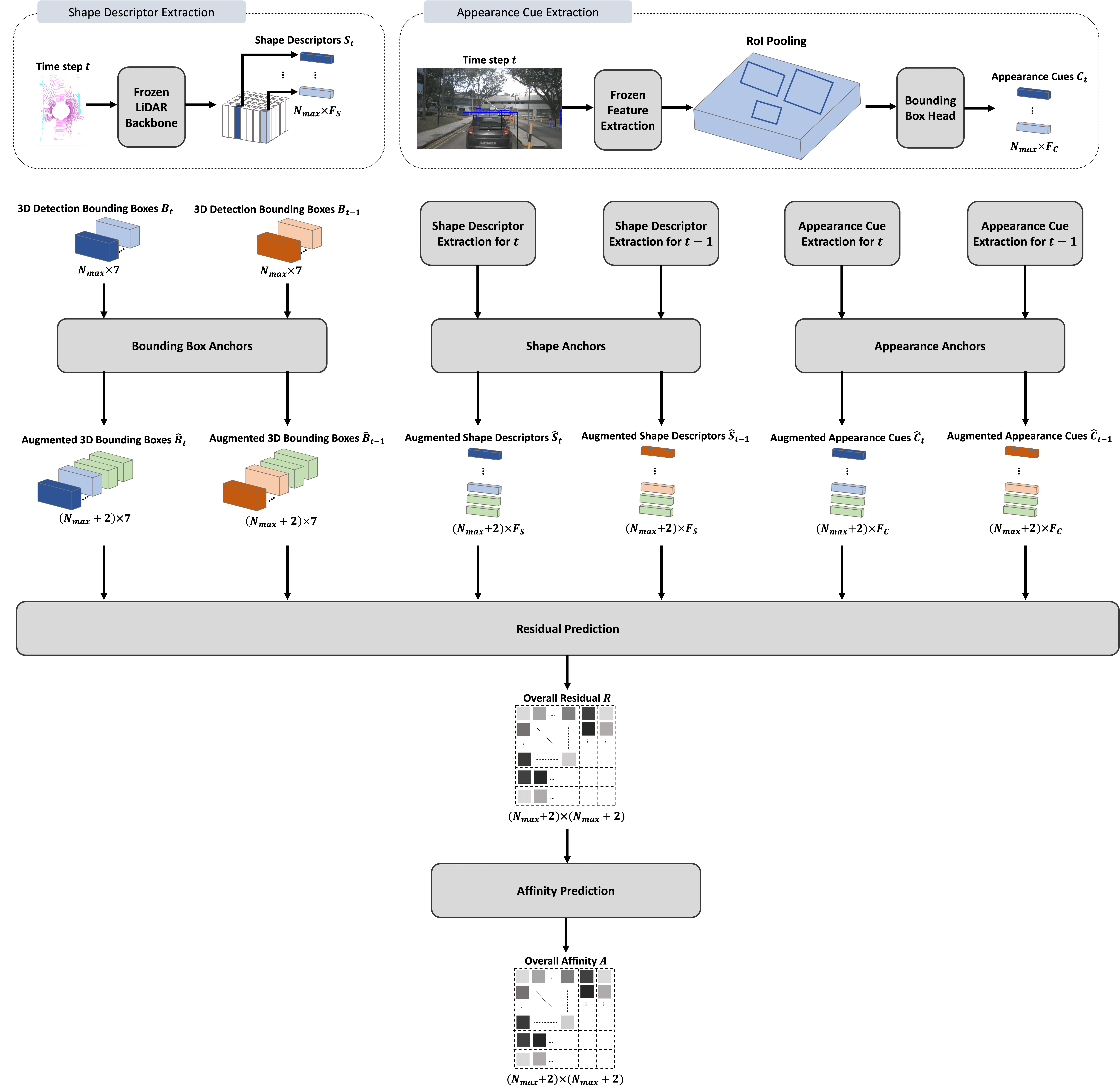}
\caption{\textbf{Algorithm Flowchart.} We first describe the two modules at the top of the figure before describing the main flowchart that utilizes these modules. In the Shape Descriptor Extraction module, we show how we extract shape descriptors from the LiDAR point cloud using the pre-trained frozen LiDAR backbone from our off-the-shelf 3D detector. This technique is also used in~\cite{sadjadpour2022shasta}. In the Appearance Cue Extraction module, we show how appearance cues are extracted for each projected 3D detection in an image. Each blue bounding box represents a projected 3D CenterPoint~\cite{yin2021center} detection in the image. A pre-trained feature extraction backbone is used from our off-the-shelf 2D detector to get the feature map, from which we do RoI pooling for each projected bounding box. Then we pass the RoI features into the Bounding Box Head, which we fine-tune to get the intermediary feature representation before the final bounding box prediction. These intermediary features are used to represent the appearance of each detected object. In the main flowchart, we use $N_{max}$ low-dimensional bounding boxes from the current and previous frames to learn bounding box representations for the FP, NB, FN, and DT anchors. The FP and NB anchors are appended to previous frame bounding boxes to form the augmented bounding boxes $\hat{B}_{t-1}$, while the FN and DT anchors are added to the current frame detections to form the augmented bounding boxes $\hat{B}_{t}$. Similarly, we extract shape descriptors and appearance cues that are then used to learn shape and appearance representations, respectively, for our anchors. Using the augmented bounding boxes, shape descriptors, and appearance cues, we find a residual that captures the spatio-temporal and shape similarities between current and previous frame detections, as well as between detections and anchors. This residual is used to predict the affinity matrix for probabilistic data associations, track lifecycle management, FP elimination, FN propagation, and track confidence refinement.}
\label{fig:pipeline}
\end{figure*}

%% file: sections/data_prep.tex
\section{Data Preparation}
Our tracking performance is evaluated using the nuScenes benchmark~\cite{caesar2020nuscenes}. This dataset is comprised of 700 training scenes, 150 validation scenes, and 150 test scenes. Each scene is 20 seconds long. Additionally, there are 7 object classes that we evaluate: bicycles, buses, cars, motorcycles, pedestrians, trailers, and trucks. 

In the nuScenes benchmark, the key frames from the camera are generated with a 2Hz frame-rate. We only use the key frames for our image data, as the input detections from CenterPoint~\cite{yin2021center} and Cascade R-CNN~\cite{cai2018cascade} are provided for the key frames only. Since the LiDAR sensor uses a sampling rate of 20Hz, we accumulate 10 LiDAR sweeps to generate the point cloud in order to match the camera sampling rate. This also makes our LiDAR point clouds become a denser 4D point cloud with an added temporal dimension. 

Moreover, we form our ground-truth affinity matrices for training using the procedure described in~\cite{sadjadpour2022shasta}. 

\input{tables/nus_test.tex}

%% file: tables/nus_test.tex
\begin{table*}[h]
\caption{\textbf{Comparison with Multimodal Trackers.} We compare our results with other multimodal detectors that use CenterPoint detections. Evaluation is on nuScenes validation set in terms of overall AMOTA and AMOTP for all object classes. The green entry is our method.}
\vspace{-10pt}
\label{table:results}
\begin{center}
\begin{tabular}{c |c c c c c}
\hline
Tracking Method & Feature Modalities & Detection Modalities & Detector(s) & AMOTA $\uparrow$ & AMOTP $\downarrow$\\
\hline 
MultimodalTracking & 2D+3D & 3D & CenterPoint & 68.7 & -\\
% \hline 
InterTrack & 2D+3D & 3D & CenterPoint & 72.1 & 0.566\\
EagerMOT & N/A & 2D+3D & CenterPoint + Cascade R-CNN & 71.2 & 0.569\\
% \hline 
\cellcolor{green!15}ShaSTA-Fuse (Ours) & \cellcolor{green!15}2D+3D & \cellcolor{green!15}2D+3D & \cellcolor{green!15}CenterPoint + Cascade R-CNN & \cellcolor{green!15}\textbf{75.6} & \cellcolor{green!15}\textbf{0.548}\\
\hline 
\end{tabular}
\end{center}
\end{table*}

%% file: sections/experiment.tex
\input{tables/ablation.tex}

\section{Experimental Results}\label{sec:experiments}

This section provides an overview of evaluation metrics, training details, comparisons against state-of-the-art 3D multi-object tracking techniques, and an ablation study to analyze our technique. 

\subsection{Evaluation Metrics}\label{sec:metrics}
The primary metrics for evaluating nuScenes are Average Multi-Object Tracking Accuracy (AMOTA) and Average Multi-Object Tracking Precision (AMOTP)~\cite{weng2019baseline}. %AMOTA measures the ability of a tracker to track objects correctly, while AMOTP measures the quality of the estimated tracks. We also present secondary metrics, including the total number of true-positives (TP), FPs, and FNs to better analyze the effectiveness of our approach for downstream decision-making tasks. We direct readers to \cite{weng2019baseline} for a more thorough definition of the metrics. 

% AMOTA measures the ability of a tracker to correctly track objects, while AMOTP measures the quality of the estimated tracks. 
AMOTA averages the recall-weighted MOTA, known as MOTAR, at $n$ evenly spaced recall thresholds. Note that MOTA is a metric that penalizes FPs, FNs, and ID switches (IDS). GT indicates the number of ground-truth tracklets. In essence, AMOTA measures the tracker's ability to correctly track objects of interest. 
{\small
\begin{align}\label{eq:amota}
    AMOTA &= \frac{1}{n-1} \sum_{r \in \{\frac{1}{n-1},\frac{1}{n-2},\dots,1\}} MOTAR \\
    MOTAR &= max\Bigg(0, 1 - \frac{IDS_r + FP_r + FN_r - (1-r)\cdot GT}{r \cdot GT}\Bigg)
\end{align}}
AMOTP averages MOTP over $n$ evenly spaced recall thresholds. For the definition of AMOTP below, $d_{i,t}$ indicates the distance error for track $i$ at time step $t$ and $TP_t$ indicates the number of true-positive (TP) matches at time step $t$. Since MOTP measures the misalignment between ground-truth and predicted bounding boxes, AMOTP is effectively measuring the quality or \textit{precision} of our estimated tracks in terms of their distance from the ground-truth tracks.
\begin{align}
    AMOTP = \frac{1}{n-1} \sum_{r \in \{\frac{1}{n-1},\frac{1}{n-2},\dots,1\}} \frac{\sum_{i,t}d_{i,t}}{\sum_t TP_t} 
\end{align}
In the nuScenes benchmark~\cite{caesar2020nuscenes}, estimated tracks within an L2 distance of 2m from ground-truth tracks are defined as TP. This margin of error is impractical for deployment in real-world driving scenarios. Thus, we emphasize AMOTP as well due to its safety implications even though the nuScenes leaderboard ranking is purely based on AMOTA.

% Additionally, we present the total number of TP, FP, and FN tracks to show that our technique optimizes for other metrics that are relevant for downstream decision-making tasks in the robot autonomy stack. 

% \input{tables/ablation.tex}

\subsection{Training Specifications}\label{sec:training}
%\noindent \textbf{Training Specifications.} 
We train a different network for each object category following~\cite{chiu2021probabilistic, stearns2022spot, sadjadpour2022shasta}. The value of $N_{max}$ depends on the object type since some classes are more common than others; this values ranges from 20 to 90 depending on the object class. We use the pre-trained VoxelNet LiDAR backbone from~\cite{yin2021center} and the pre-trained Cascade R-CNN detection~\cite{cai2018cascade}. The LiDAR backbone is frozen, and the Cascade R-CNN is frozen except for the Bounding Box Head that we fine-tune for this task. Additionally, due to the class imbalance between FPs and TPs, we downsample the number of FP detections during training.
Finally, we fix the thresholds to $\tau_{fp}=0.7$ for FP elimination, $\tau_{fn}=0.5$ for FN propagation, $\tau_{nb}=0.5$ for NB initialization, and $\tau_{dt}=0.5$ for DT termination across all object classes. For $\tau_{iou}$, we use $0.4$ for small objects (bicycle, motorcycle, pedestrian), 0.6 for cars and trucks, and 0.9 for buses and trailers.

\subsection{Comparison with State-of-the-Art Multimodal Tracking} Our nuScenes validation results can be seen in Table~\ref{table:results}. Evidently, our technique gives significant boosts for tracking accuracy, while retaining the best precision. Thus, ShaSTA-Fuse is effective in identifying correct tracks, while ensuring precise track localization for increased safety.

\subsection{Ablation Studies}
In this section, we conduct ablative analysis on our tracking algorithm to assess the benefits of camera-LiDAR sensor fusion. In Table~\ref{table:fusion_ablation}, we examine the effects of adding different types of camera information. It is evident that fusing LiDAR and camera information through combining 2D and 3D detectors yields far more gains in accuracy than extracting appearance cues from the camera images in a neural network. This is consistent with results we have seen in past literature on this topic, where fusing LiDAR and camera features in a neural network can yield as little as less than 0.1\% improvement in accuracy for each object~\cite{wang2022camo, chiu2021probabilistic}, while bounding box fusion yields significant gains particularly for objects like bicycles~\cite{wang2022deepfusionmot, kim2021eagermot}. 

%% file: tables/ablation.tex
\begin{table*}[h]
\caption{\textbf{Ablation on Camera-LiDAR Fusion.} We assess the effect of different camera-LiDAR fusion techniques on tracking accuracy. First, we begin with LiDAR-only shape and spatio-temporal information without any form of confidence refinement. In the second row, we fuse appearance features extracted from the camera's image data with the LiDAR information from the previous row without any confidence refinement. Then, in the third row, we have the fused camera-LiDAR features but only use information from the LiDAR-based 3D detections for the confidence refinement. Finally, in the last row we integrate 2D and 3D detection information for the full multimodal confidence refinement with multimodal feature extraction.  Evaluation is on the nuScenes validation set in terms of overall AMOTA for a subset of object classes. The green entry is our full method. }
\vspace{-10pt}
\label{table:fusion_ablation}
\begin{center}
\begin{tabular}{c |c c c c c c }
\hline
Method & Detector(s) & Bicycle & Car & Motorcycle & Pedestrian \\
\hline 
LiDAR-only w/o Confidence Refinement & CenterPoint & 49.4 & 85.3 & 70.1 & 80.9\\
% \hline
+ Appearance Residual & CenterPoint & 49.8 & 85.3 & 71.7 & 80.9\\
% \hline 
+ LiDAR-only Confidence Refinement & CenterPoint & 57.7 & \textbf{85.7} & 73.6 & 81.4\\
% \hline
\cellcolor{green!15}+ Multimodal Confidence Refinement & \cellcolor{green!15}CenterPoint + Cascade R-CNN & \cellcolor{green!15}\textbf{72.3} & \cellcolor{green!15}85.6 & \cellcolor{green!15}\textbf{78.4} & \cellcolor{green!15}\textbf{82.8}\\
\hline
\end{tabular}
\end{center}
\end{table*}

%% file: sections/conclusion.tex
\section{Conclusion}
% \textcolor{red}{TODO}

In summary, ShaSTA-Fuse extends the state-of-the-art affinity-based 3D MOT framework to accommodate multimodal camera-LiDAR tracking. This work explores pertinent questions about the benefits of camera-LiDAR sensor fusion, particularly the tradeoff between accuracy and computational efficiency. Based on our experimental results and the outcomes of past works in this field, we have found that the inexpensive fusion of abstracted camera-based 2D detections and LiDAR-based 3D detections is far more computationally efficient, while yielding significantly higher gains in accuracy. Thus, we encourage the community to focus more on developing such techniques in this under-explored aspect of multimodal fusion. 